\documentclass[final]{cvpr}

\usepackage{times}
\usepackage{epsfig}
\usepackage{graphicx}
\usepackage{amsmath}
\usepackage{amssymb}
\usepackage{subfigure}

\usepackage[pagebackref=true,breaklinks=true,colorlinks,bookmarks=false]{hyperref}

\def\cvprPaperID{****} % *** Enter the CVPR Paper ID here
\def\confYear{CVPR 2021}

\begin{document}

%%%%%%%%% TITLE
\title{
   $AIR^2$ for Interaction Prediction
}

\author{David Wu\\
\and
Yunnan Wu\\
}

\maketitle

\begin{figure*}
   \begin{center}
   \includegraphics[width=\textwidth] {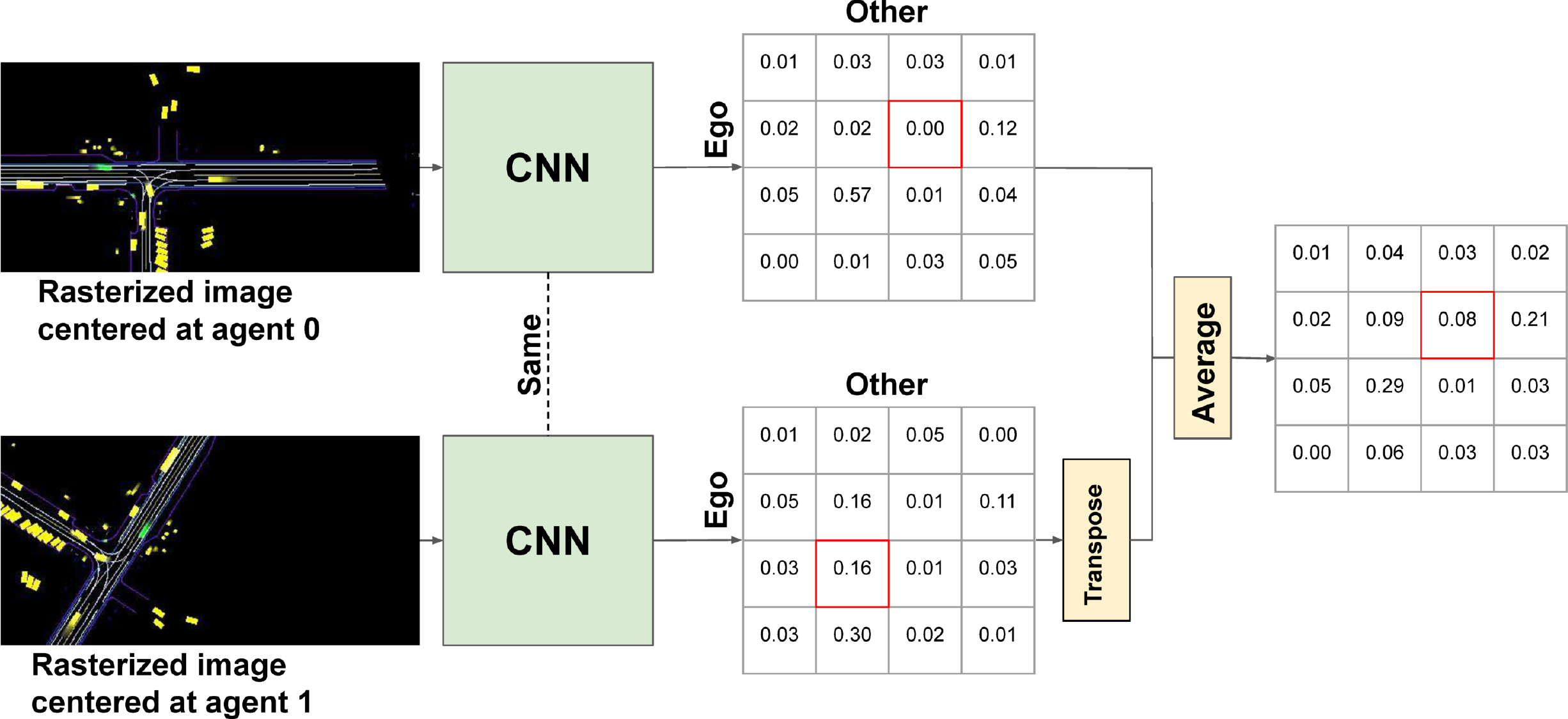}
   \end{center}
      \caption{Illustration of $AIR^2$ for Interaction Prediction. We apply the same CNN to rasterized images centered at each agent to predict joint confidences over the cartesian product anchors. Then the two predictions are ensembled. The red box illustrates a potential ground truth assignment.}
   \label{fig:confidences}
   \end{figure*}
%%%%%%%%% ABSTRACT
\begin{abstract}
   The 2021 Waymo Interaction Prediction Challenge introduced a problem
   of predicting the future trajectories and confidences of two interacting agents jointly. We developed a solution that takes an anchored marginal motion prediction model with rasterization and augments it to model agent interaction. We do this by predicting the joint confidences using a rasterized image that highlights the ego agent and the interacting agent. Our solution operates on the cartesian product space of the anchors; hence the $``^2"$ in $AIR^2$. 
   Our model achieved the highest mAP (the primary metric) on the leaderboard.
\end{abstract}

%%%%%%%%% BODY TEXT
\section{Introduction}

The problem of interaction prediction, stated clearly, is the problem of predicting the future states of two chosen agents with interesting interactions jointly, meaning to predict where both agents will be at each time step. It is a relatively new problem, and a simple approach would be to take a model trained for the classical problem of Motion Prediction, run it on both agents, then do a simple cartesian product between the two predicted sets of future trajectories and confidences, taking the pairs which have highest multiplied confidences.

Such a model achieves poor results, but these results are not unexpected. This is because the simple model is essentially treating the actions of the two agents as independent of each other, while in reality the movement of one agent will depend heavily on the movement of the other. What this problem requires is a model that is aware of the interaction between the two agents, as well as how this will affect their future trajectories. 

In this paper, we describe our solution, 
$AIR^2$, used for the interaction prediction task, and describe techniques that can help the model become more aware of interactions and make better predictions. Our Interaction Prediction model $AIR^2$ builds upon our Motion Prediction model, AIR. However, the proposed interaction prediction solution is generally compatible with anchored motion prediction models with rasterization.

The key features of our solution are as follows:
\begin{enumerate}
   \item We make extensive use of the idea of anchoring, introduced in MultiPath \cite{ChaiSBA2003} for motion prediction. Our marginal model predicts multiple modes, each with a trajectory and confidence, anchored to one representative pattern. This decomposes the problem into two parts: intent classification, and regression given intent. Anchoring introduces a clean division of responsibilities.
   \item We develop a framework to build upon a good marginal model by adding a joint-confidence predictor out of the marginal model intermediate embeddings. We introduce interaction modeling in the carteisan product space formed by the anchored multi-mode predictions, hence the $``^2"$ in $AIR^2$. For the current solution, we took a simple cartesian product of the trajectories predicted by the marginal model; only the joint confidences are explicitly modeled with interaction modeling.
   \item We deal with significant imbalance of data by sampling from vehicle data, pedestrian data, and cyclist data separately. In addition, we use expert subnetworks gated by object types in our models, so that a single model can handle all of the data, and specialize in different object types to capture their unique behaviors and characteristics. This reduces the competition in parameter learning driven by different object types.
   \item We introduce an empirical method, called \emph{metric-loss sensitivity analysis}, which aims at estimating how a small change in each component of the loss affects the final metric (mAP). We use this method to set various loss weight parameters.
\end{enumerate}
%-------------------------------------------------------------------------

\subsection{Related Work}
 Due to time constraints, we did not extensively search for literature regarding Interaction Prediction. 
However, aspects of our Motion Prediction approach were inspired by the first place \cite{Lyft1st} and third place \cite{Lyft3rd} winners in the Lyft L5 Motion Prediction challenge, as we also pursued a CNN backbone + predicting heads structure after rasterization. In addition, MultiPath \cite{ChaiSBA2003} introduced the idea of anchoring for Motion Prediction, which is a crucial part of our solution.

\section{$AIR$ for Motion Prediction}
While the focus of this paper is on Interaction Prediction, the problem of Interaction Prediction is closely intertwined with the problem of Motion Prediction. For the sake of completeness, we describe our Motion Prediction model (Our $AIR$ Model for Motion Prediction ranked 5th place on the Waymo Motion Prediction challenge leaderboard), but we also remark that there are other Motion Prediction models that would also fit into our framework for Interaction Prediction.

\begin{figure}%[ht!]
   \centering
   \includegraphics[width=\linewidth]{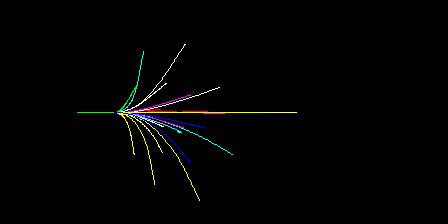}
   \caption{Cyclist clusters\label{fig:clusters}}
   \end{figure}

Why is our Motion Prediction model named $AIR$? ``AIR" stands for three things that are vitally important to the structure of our model:
\begin{enumerate}
   \item $\mathbf{A}$ stands for \emph{anchored}, meaning that our trajectories are split into different modes and that our model makes one prediction for each mode, and predictions are also supervised based on the mode assignment of the ground truth, following the MultiPath paper \cite{ChaiSBA2003}. The anchors are obtained by clustering ground truth trajectories in the training data using K-Means clustering, see, e.g., Figure~\ref*{fig:clusters}. 
   \item $\mathbf{I}$  stands for \emph{interpolated}, meaning that our model predicts 8 control points, instead of full trajectories, which are then interpolated to get the full 80 time steps prediction. The intuition behind splining control points instead of predicting all 80 time steps directly is the human knowledge that trajectories will be smooth. Using splining helps us encode that knowledge into the model, and avoids wavy, unrealistic, trajectories. We implemented this using tensorflow graphics.
   \item $\mathbf{R}$ stands for \emph{rasterization}. Our model learns crucial information about a scenario through a rasterized image, which provides it with a basic understanding of the scenario that helps it make a prediction. Rasterization, or “drawing” information about a scene onto an image that a model will take in to make predictions for a certain agent, is a standard technique in the field of Motion Prediction. The pictures generated via rasterization are human interpretable, and are an excellent way to condense all relevant information about a scenario in a single, usable, input.
\end{enumerate}

\begin{figure}[t]
   \begin{center}
   \includegraphics[width=\linewidth]{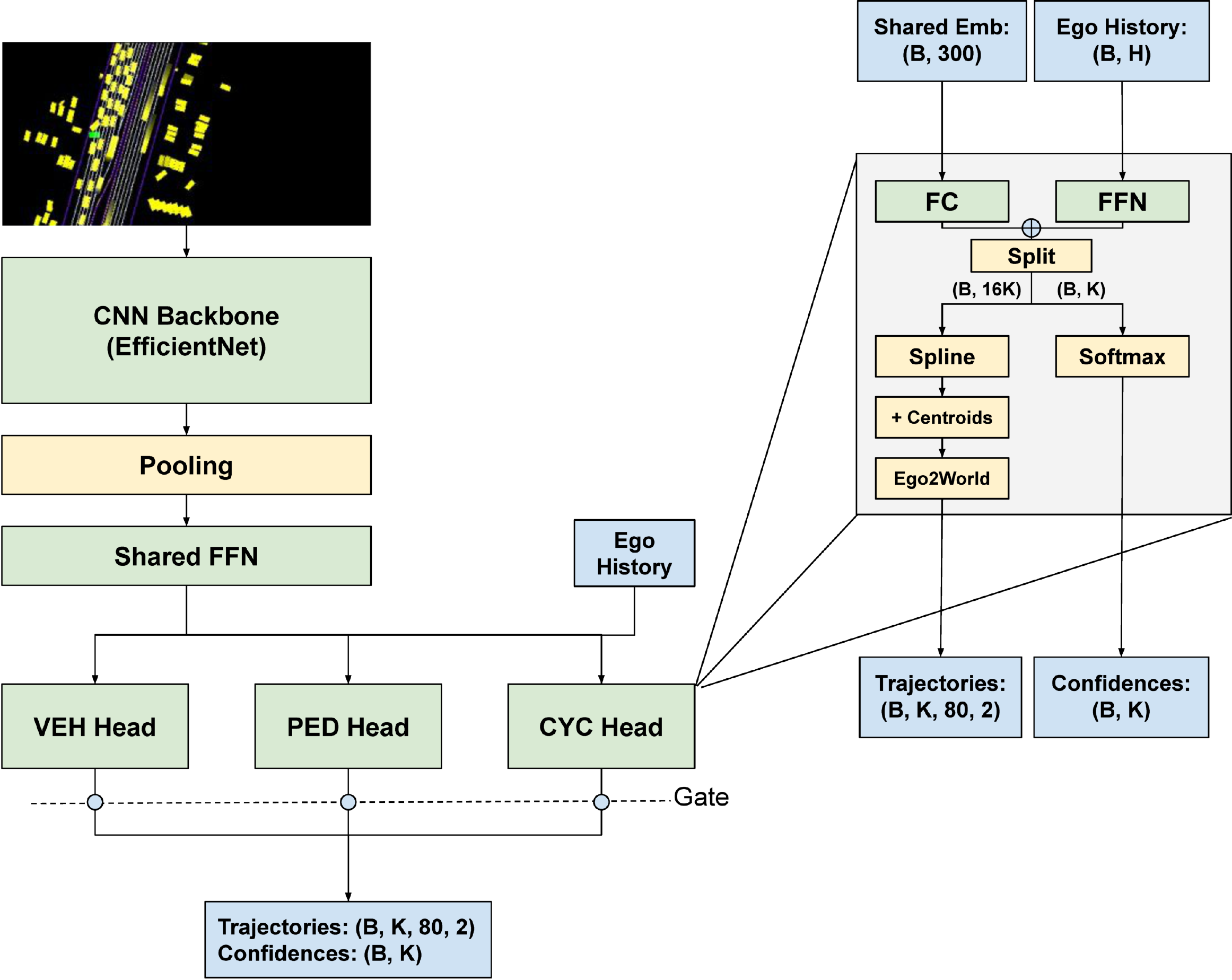}
   \end{center}
      \caption{AIR Model Architecture}
   \label{fig:AIR}
   \vspace{-.2in}
   \end{figure}

Our Motion Prediction model, AIR (See Figure \ref*{fig:AIR}), takes in a 224x448x3 image rendered in ego coordinates as input, then runs it through a pretrained CNN backbone and average pooling to generate embeddings. These embeddings then go through a shared FFN and then go to per object type expert heads for predictions. Inside a head, past states for the ego agent are run through an FFN to create embeddings carrying information about the past states. These are then added to the FC output of the shared embeddings before being split into control points and confidences. The control points are first splined, then the centroid for each anchor is added on, so that the model is predicting a residual for each anchor. Then, the final predictions are transformed into world coordinates, as the model has been working in ego coordinates up until now. The confidences are softmaxed. Then, final predictions are gated by object type and sent off to the loss for back propagation.

\begin{figure*}
   \begin{center}
   \includegraphics[width=\textwidth] {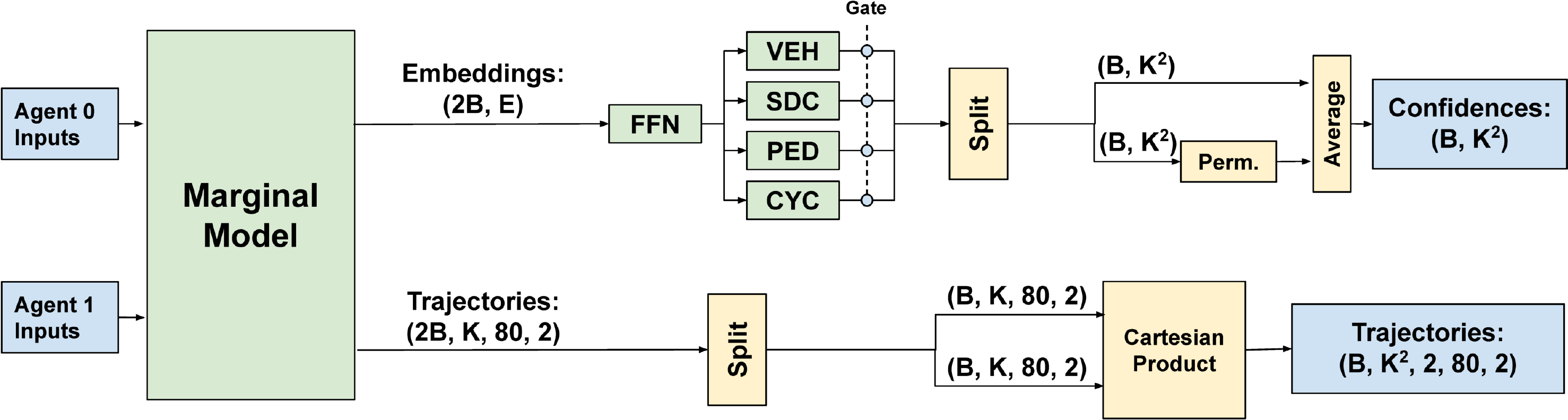}
   \end{center}
      \caption{$AIR^2$ for Interaction Prediction. $K$ is the number of anchors (32 in our experiments). $B$ is the batch size.}
   \label{fig:arch}
   \end{figure*}
\section{$AIR^2$ for Interaction Prediction}

The core task of Interaction Prediction is to predict the joint  probability of two agents, whereas models for the Motion Prediction task are more concerned about predicting the marginal probabilities. It would be great if we could leverage a good Motion Prediction model and build on top of it, with considerations for joint interaction prediction. 
Figure~\ref*{fig:confidences} illustrates the key idea of our approach. We augment the marginal model with extra FFNs that predict joint probabilities over the cartesian product anchors. This process is done once for agent 0 and once for agent 1, resulting in two predictions of the joint probability matrix. After transposing the second predicted matrix, we then average the two matrices, resulting in a ``natural ensemble'' of the joint probabilities. For each training example, the ground truth is assigned to a pair of anchor trajectories, corresponding to one cell in the grid. This triggers back propagation for Interaction Prediction.

\subsection{Model Architecture}
The $AIR^2$ model architecture is illustrated in Figure~\ref*{fig:arch}.
Our model builds upon a given marginal Motion Prediction model, which outputs multiple anchored trajectories with confidences. We operate in the product anchor space and use the same clusters as the marginal model. In addition, the marginal model also outputs some intermediate embeddings. These embeddings are then run through a shared FFN (with ReLU activation at the end) and through separate specialized object type heads. The outputs from these heads are then gated depending on the true object type of the example. We used separate joint confidence predicting FFNs  for vehicle, pedestrian, and cyclist, as well as an extra FFN for self-driving cars. The intuition behind splitting our FFN into different FFNs for different object types is that different object types have different behaviors. For example, vehicles are more likely to let other agents go, while pedestrians are more likely to go ahead. We also perform clustering for each object type separately, due to a belief that different object types have different overall intents and behaviors. 

Then, the outputs are split, permuted, and averaged to give the final confidences. The permutation is needed because agent 0 predicts a confidence grid $\{p_{ij}\}$ from their perspective, while agent 1 predicts a confidence grid $\{p_{ji}\}$ from their perspective, and the two must match.
As illustrated by the bottom of Figure \ref{fig:arch}, the trajectories are just split, and then go through a simple cartesian product to give the final trajectories. 

\subsection{Interaction Loss}

Our loss was inspired by \cite{ChaiSBA2003}. This loss is calculated by first calculating anchor assignments $i^*$ and $j^*$ for the ground truth trajectories using our precalculated clusters (See Figure \ref{fig:clusters}), where $i^*$ refers to the assignment for the first agent, and $j^*$ refers to the assignment of the second. We then use the following calculations:

\begin{equation}
\begin{aligned} 
L &= w_{reg} * L_{reg} + w_{cls} * L_{cls}
\end{aligned}
\end{equation}
\begin{equation}
   L_{cls} = -\log{p_{i^*j^*}} + w_{m} * L_{m}
\end{equation}
\begin{equation}
   L_{m} = -\log{\sum_{j}p_{i^*j}} - \log{\sum_{i}p_{ij^*}}
\label{eq:L_m}
\end{equation}

\begin{equation}
L_{reg} = L_{reg^{(0)}} + L_{reg^{(1)}}
\end{equation}

\begin{equation}
L_{reg^{(0)}} = \frac{1}{2}\sum_{t=1}^{80}{
\left(   (\hat{x}_{t}^{(i^*)} - x_{t}^{(0)})^2 + 
   (\hat{y}_{t}^{(i^*)} - y_{t}^{(0)})^2 \right)   
}
\end{equation}

\begin{equation}
   L_{reg^{(1)}} = \frac{1}{2}\sum_{t=1}^{80}{
   \left(   (\hat{x}_{t}^{(j^*)} - x_{t}^{(1)})^2 + 
      (\hat{y}_{t}^{(j^*)} - y_{t}^{(1)})^2 \right)   
   } 
   \end{equation}

Due to concerns about confidences becoming diluted over too many modes and not enough supervision for each mode, our loss is modified to supervise both of the single agent cluster prediction assignments for each example, in addition to the usual intersection of the two single agent cluster assignments. This corresponds to the $L_m$ term (\ref{eq:L_m}). We introduce an marginal loss weight $w_m$ to allow experimenting with this modification.

The loss formula is a weighted sum of the classification loss and the regression loss. How should we set their relative weights? One approach is to perform hyperparameter tuning by experimenting with many settings, which incurs heavy computational resource usage.
Instead of doing that, we introduce an empirical method for weighting the classifcation and regression losses, which we call \emph{metric-loss sensitivity analysis}.
The high level idea is to measure the mAP gain resulting from a unit of decrease in classification loss vs a unit of decrease in regression loss, and then use such sensitivity information to set the loss weights.

The process works as follows: We partially reveal to the model the correct trajectories or confidences using the following equations:
\begin{equation}
   \begin{aligned}
   \mathbf{\hat{p}} &= (1 - \alpha)*\mathbf{p} + \alpha*\mathbf{p_{gt}}\\
   \mathbf{\hat{x}} &= (1 - \alpha)*\mathbf{x} + \alpha*\mathbf{x_{gt}},
   \end{aligned}
   \end{equation}
where $\alpha$ is a small positive number (e.g., 0.1), $\mathbf{p}$ refers to the confidence vectors, and $\mathbf{x}$ refers to the trajectories. Then, we measure the decrease in classification loss when we reveal the confidences, and the decrease in regression loss when we reveal the trajectories, and measure the corresonding mAPs. Then, we weight our classification and regression losses so that we generate a similar amount of mAP for the same decrease in classification versus regression loss. The following is a detailed explanation of the mechanics of the approach.

Suppose $\alpha=0.1$. Consider a given model, and consider one training example. For this example, suppose the ground truth anchor index is $k$. We compute the ground truth confidence vector $\mathbf{p_{gt}}$ as the one-hot vector of index $k$. Suppose that the model predicts a confidence vector $\mathbf{p}$. To measure the sensitivity of mAP to the classification loss, we replace our predicted confidences with new confidences, calculated by the following formula: $\mathbf{\hat{p}} = 0.9*\mathbf{p} + 0.1*\mathbf{p_{gt}}$. This typically would reduce classification loss and improve mAP. With this change in confidences, we calculate the change in mAP and the change in classification loss; the ratio reflects how much mAP improvement results in a unit decrease of classification loss.

Similarly, we can compute the sensitivity of mAP to the regression loss.
For one example with ground truth anchor index $k$, suppose that the model currently predicts trajectories $\mathbf{x}$ for anchor $k$, and the ground truth trajectory is $\mathbf{x_{gt}}$. To measure the sensitivity of mAP to the regression loss, we replace our predicted trajectories with new trajectories, calculated by the following formula: $\mathbf{\hat{x}} = 0.9*\mathbf{x} + 0.1*\mathbf{x_{gt}}$, for anchor $k$. We then leave the other anchors' trajectories unchanged. This typically would reduce regression loss and improve mAP. With this change in trajectories, we calculate the change in mAP and the change in regression loss; the ratio reflects how much mAP improvement results in a unit decrease of regression loss.

After that, we calculate our regression and classification weights, making them proportional to the sensitivity ratios computed above.
Doing so will result in similar amounts of mAP for the same decrease in loss between classification and regression. This helps align the loss optimization with mAP maximization.

\subsection{Input Representations}
\begin{figure}%[ht!]
   \centering
   \subfigure{\includegraphics[width=\linewidth]{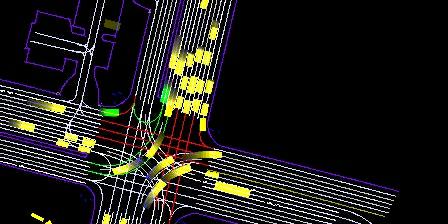}}
   \subfigure{\includegraphics[width=\linewidth]{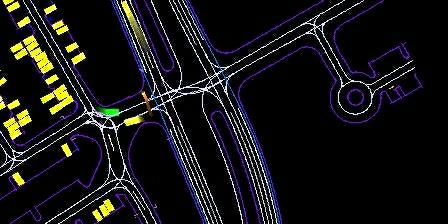}}
   \caption{(a) Example rasterization with both agents rendered in green.\label{fig:green} (b) Example of rerasterization \label{fig:RR}}
   \end{figure}
We introduce two forms of input representations:
\begin{enumerate}
   \item The first type of images that we use are images with one agent, the ego agent, at (112, 112), colored in green, and the other interacting agent also colored in green. This tells the model what agent is the interacting agent, and lets it use all information in the scene in order to model interactions. (See Figure \ref{fig:green}(a))
   \item The second type of images that we use are rerasterized images, which provide the model with an idea of the movements of the other agent. Specifically, our model takes the prediction output by a shared Motion Prediction model for each agent and rasterizes it onto images for the other agent, then feeding the new image as input to our interaction model (See Figure \ref{fig:RR} (b)).
\end{enumerate}
For our submissions, we used rerasterized images to train our model.

\subsection{Dealing with Imbalanced Training Data}
Due to an imbalance in the amount of vehicle, cyclist, and pedestrian data, our models were trained with a custom dataset, consisting of examples sampled from three separate vehicle, cyclist, and pedestrian datasets with equal probability. We implemented this in a few lines using $\tt{tf.data.experimental.sample\_from\_datasets}$.

\section{Experiments}
\begin{table}
   \begin{center}
   \begin{tabular}{|l|c|}
   \hline
   Model & mAP\\
   \hline\hline
   $AIR^2$, our model, ensembled & \bf{0.0963}   \\
   King Crimson & 0.0897 \\
   HeatIRm4 & 0.0844 \\
   Waymo LSTM Baseline & 0.0524\\
   \hline
   \end{tabular}
   \end{center}
   \caption{mAP Results on Leaderboard. \label{table:mAP}}
   \end{table}
Our model was trained on the Waymo Open Dataset, collected by Waymo’s autonomous vehicles. Our images are generated using rasterization(See Figure \ref{fig:green}), with one image for each agent, centered on the agent. Images are 224x448x3 with standard RGB channels. We implemented our solution using Tensorflow. The mAP results are given in Table~\ref*{table:mAP}.

{\bf Clustering:} We cluster the ground truth trajectories (x and y coordinates) from the Waymo Motion Prediction dataset training data with a modified version of K-Means. 
Our modified K-Means algorithm alternates between two steps:
\begin{enumerate}
   \item First, the algorithm calculates the distance between each of the points and the existing centroids, and assigns each point to the nearest cluster. The distance calculations use the availability as the mask. 
   \item Then, the centroids are updated to be the average of all of the points that were assigned to the specific centroid. We average over the number of available points for each time step.
\end{enumerate}

Clustering was done separately for vehicles, pedestrians, and cyclists. For vehicles, we use 32 clusters; for pedestrians, we use 8 clusters; for cyclists, we use 30 clusters. This results in $K=32$ anchors for motion prediction and $K*K=1024$ anchors for interaction prediction. Padding is used for pedestrian and cyclist predictions.

During the clustering, we discovered that there was corruption in the data, that would disrupt our clustering. Specifically, we found that around 0.2 percent of the ground truth trajectories had sudden impossible changes in position, whereas the validity mask indicated that the time steps had valid data. We handled this by filtering out the ground truth trajectories that had too big of a change in position between consecutive time steps.

{\bf Spline:} For splining in the $AIR$ model, we use cardinal cubic polynomial bspline. This was implemented using $\tt{tensorflow\_graphics.math.interpolation.bspline}$. 

{\bf CNN Backbone:} For the AIR model, we used the EfficientNetB3 as the CNN backbone, which was pretrained on ImageNet.

{\bf Loss Weights:} In our experiments for $AIR^2$ for interaction prediction, $w_{cls}=60$ and $w_{reg}=1$.

For our submissions, we trained our model with 1000 training files, as well as the first 100 validation files. In addition, we ensembled 4 models with different runs of the same config (2 of them are different snapshots of the same training run, and the other 2 are different runs) with simple averaging. The 4 models all started with the same marginal AIR model. Our submissions used re-rasterized input images and a marginal loss weight of $w_m=1$. For the source code and the detailed parameter configs, please refer to our github repo at \url{https://github.com/david9dragon9/AIR}. Due to time constraints, we have not finished the ablation studies. The results will be reported in our github repo later.

\section{Conclusion and Future Work}
We developed techniques for modeling the interaction between two agents in a scenario. Our interaction prediction model builds upon an anchored marginal motion prediction model with rasterization, e.g., the $AIR$ model for motion prediction. We augment the marginal model to model agent interaction by predicting the joint confidences with a rasterized image that highlights the ego agent and the interacting agent.

Our current $AIR^2$ model explicitly models the joint confidences but takes a simple treatment for the trajectories, by directly outputting the cartesian product of the trajectories. In the future, we could explore techniques for modifying the trajectories predicted by the Motion Prediction model to better take interactions into account.

The current Interaction Prediction problem formulation considers two interacting agents. In real world, each scene has many agents that are interacting together. We are interested in exploring how to take into account all of these interactions to arrive at more accurate understanding of the environment as a whole. 

%------------------------------------------------------------------------

{\small
\bibliographystyle{ieee_fullname}
\bibliography{egbib}
}

\end{document}